\documentclass[lettersize,journal]{IEEEtran}
\usepackage{tikz,xcolor}
\usepackage[implicit=false]{hyperref}
\hypersetup{hidelinks,
	colorlinks=true,
	allcolors=black,
	pdfstartview=Fit,
	breaklinks=true}
\usepackage{amsmath,amsfonts}
\usepackage{algorithmic}
\usepackage{algorithm}
\usepackage{array}
\usepackage[caption=false,font=normalsize,labelfont=sf,textfont=sf]{subfig}
\usepackage{textcomp}
\usepackage{stfloats}
\usepackage{url}
\usepackage{verbatim}
\usepackage{graphicx}
\usepackage{cite}
\usepackage{booktabs}
\usepackage{multirow}
\usepackage{amssymb}
\usepackage{pifont}
\usepackage{orcidlink}
\begin{document}

\title{TAVP: Task-Adaptive Visual Prompt for Cross-domain Few-shot Segmentation}

\author{Jiaqi Yang, Yaning Zhang, Jingxi Hu, Xiangjian He\orcidlink{0000-0001-8962-540X}, \IEEEmembership{Senior Member,~IEEE}, Linlin Shen\orcidlink{0000-0003-1420-0815}, \IEEEmembership{Senior Member,~IEEE}, Guoping Qiu\orcidlink{0000-0002-5877-5648}, \IEEEmembership{Senior Member,~IEEE}
\thanks{This work is partially supported by the National Natural Science Foundation of China under Grant (82261138629) and (12326610); Guangdong Provincial Key Laboratory under Grant 2023B (1212060076);  the NSFC project (UNNC Project ID B0166); Yongjiang Technology Innovation Project (2022A-097-G); and Shenzhen Municipal Science and Technology Innovation Council under Grant JCYJ (20220531101412030).
\textit{(Corresponding author: Xiangjian He)}}
\thanks{\quad Jiaqi Yang, Jingxi Hu, Xiangjian He and Guoping Qiu are with School of Computer Science, University of Nottingham Ningbo China, Ningbo, Zhejiang, China(e-mail: jiaqi.yang2@nottingham.edu.cn; jingxi.hu@nottingham.edu.cn; sean.he@nottingham.edu.cn; guoping.qiu@nottingham.edu.cn).}
\thanks{\quad Yaning Zhang is with the Faculty of Computer Science and Technology, Qilu University of Technology (Shandong Academy of Sciences), Jinan, China(e-mail: zhangyaning0321@163.com).}
\thanks{\quad Linlin Shen is with Computer Vision Institute, School of Computer Science \& Software Engineering, Shen Zhen, Guang Dong, China(e-mail:llshen@szu.edu.cn).}
}
% \markboth{Journal of \LaTeX\ Class Files,~Vol.~14, No.~8, December~2024}%


\maketitle

\begin{abstract}
While large visual models (LVM) demonstrated significant potential in image understanding, due to the application of large-scale pre-training, the Segment Anything Model (SAM) has also achieved great success in the field of image segmentation, supporting flexible interactive cues and strong learning capabilities. However, SAM's performance often falls short in cross-domain and few-shot applications. Previous work has performed poorly in transferring prior knowledge from base models to new applications. To tackle this issue, we propose a task-adaptive auto-visual prompt framework, a new paradigm for Cross-dominan Few-shot segmentation (CD-FSS). First, a Multi-level Feature Fusion (MFF) was used for integrated feature extraction as prior knowledge. Besides, we incorporate a Class Domain Task-Adaptive Auto-Prompt (CDTAP) module to enable class-domain agnostic feature extraction and generate high-quality, learnable visual prompts. This significant advancement uses a unique generative approach to prompts alongside a comprehensive model structure and specialized prototype computation. While ensuring that the prior knowledge of SAM is not discarded, the new branch disentangles category and domain information through prototypes, guiding it in adapting the CD-FSS. Comprehensive experiments across four cross-domain datasets demonstrate that our model outperforms the state-of-the-art CD-FSS approach, achieving an average accuracy improvement of 1.3\% in the 1-shot setting and 11.76\% in the 5-shot setting.
\end{abstract}


\section{Introduction}
\label{sec1}
\IEEEPARstart{T}{raditional} deep networks relied heavily on extensive annotated data to achieve high precision performance \cite{chen2014semantic}. However, data annotation is a time-consuming task that requires substantial human resources, particularly for intensive pixel-level annotation tasks such as medical image and remote sensing image segmentation. Therefore, Few-shot semantic segmentation (FSS) was introduced to narrow this gap \cite{wang2019panet}, aiming to reduce the need for labelling. Besides, most of the few-shot learning (FL) methods mainly focus on learning the relationship between support and query within the same domain, which always requires fine-tuning on the target domain \cite{TIAN2024110091,10145054}. The high-level features extracted are class-agnostic but lack domain generalization, meaning that while FL methods are capable of generating new categories, they perform poorly in generating new domains. The previous methods have limitations in that input-space-based enhancement requires expert knowledge to design enhancement functions, while feature-based enhancement usually relies on complex adversarial training \cite{guan2021domain}. Thus, Cross-domain Few-shot Segmentation (CD-FSS) \cite{lei2022cross} came up for solving segmentation tasks on medical, remote sensing, and other images. 

However, previous deep models may lead to poor generalization on unseen out-of-domain data, which limits their use in Cross-domain Few-shot scenarios. Recently, Large fundamental visual models (LVM) have made significant progress in natural image segmentation \cite{ji2023sam,he2023accuracy,zhou2023can}, including medical \cite{zhang2023method} and remote sensing \cite{zhang2023iterative} image segmentation. The Segment Anything Model (SAM) \cite{kirillov2023segment} was trained with over one billion masks and achieved unprecedented generalization capabilities on natural images. Additionally, some research has shown that proper adjustments to SAM can be applied in medical image segmentation \cite{zhang2023method} and zero-shot tasks. These advances suggest that powerful segmentation models with generalization capabilities can be used without designing complex networks due to time-consuming retraining. Some early works have used pre-trained models on natural or medical images and achieved good performance \cite{kirillov2023segment,tang2023can}. However, due to the inflexible capacity of pre-trained models and extensive few-shot methods such as disentanglement domain classifier \cite{fu2021meta}, the cross-domain generalization ability of deep models has not been effectively improved.

\begin{figure*}[!t]
\includegraphics[width=7.1in, height=4in]{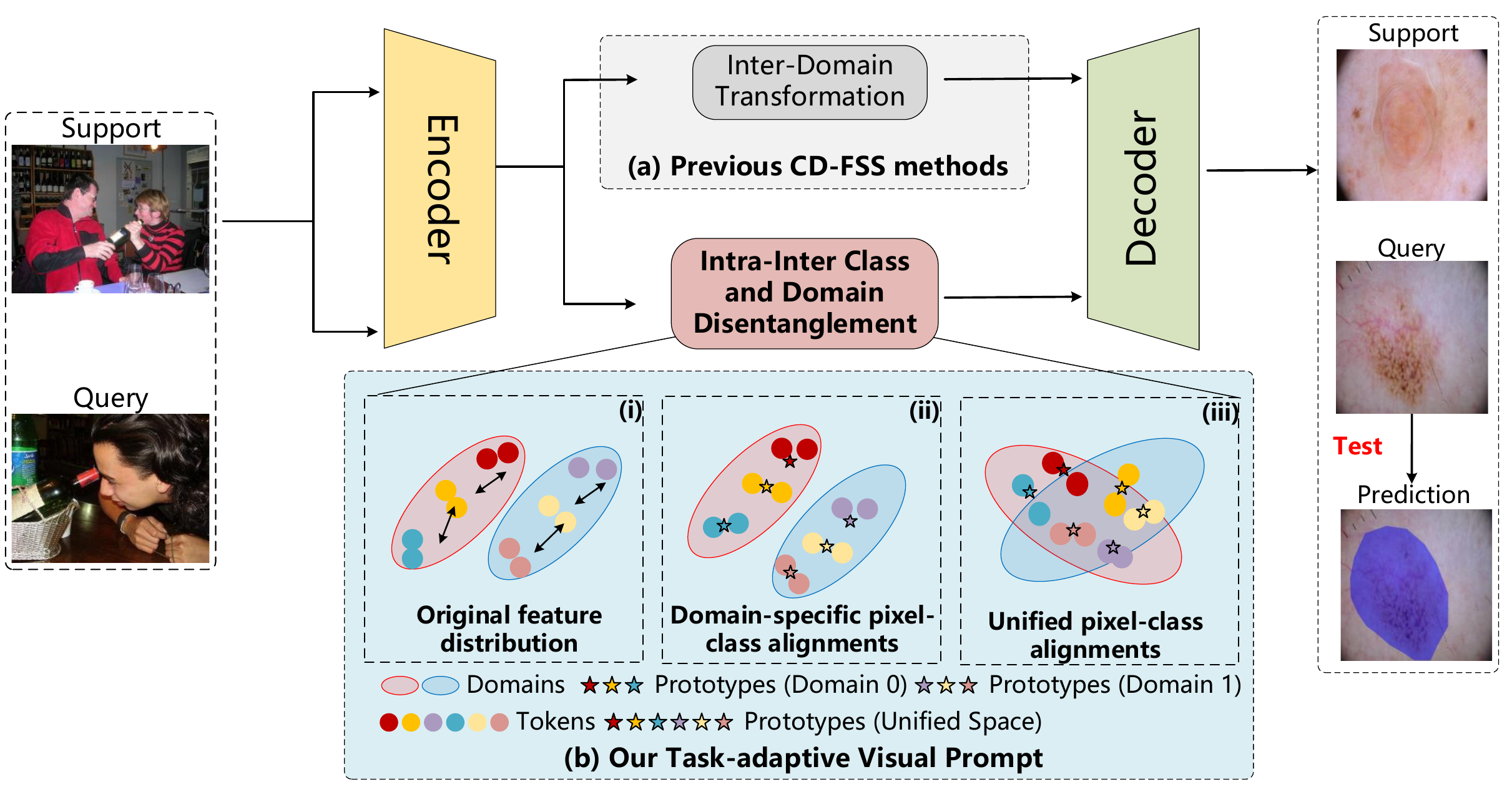}
    \caption{(a) The existing cross-domain segmentation method based on SAM. (b) Our Task-adaptive Visual Prompt. (i) There is no interaction between features from different categories in the original feature distribution. (ii) In a specific domain, prototypes are used to make semantic distinctions between categories to achieve clustering. (iii) Finally, inter-category distinction and intra-category strong constraints are achieved in a unified space. }
    \label{fig: comparison with popular paradigm}
\end{figure*}

While these LVM-based methods have enhanced model performance in certain professional domains, several limitations remain, which can be summarized as follows: \textbf{(1) Poor Generalization to Specific Domains:} While LVM-based methods enhance performance in certain professional fields, SAM struggles to generalize effectively to domain-specific tasks. \textbf{(2) High Cost of Domain-Specific Adaptation:} Adapting SAM to specific domains involves significant costs, including data collection, sample labelling, and model training. This dependency on large-scale domain-specific datasets makes the approach resource-intensive and impractical for domains with limited annotated data. \textbf{(3) Difficulty in Covering All Domains:} It is infeasible to exhaustively enumerate and address all possible specific domains. This inherent limitation restricts the scalability of SAM-based approaches in scenarios where diverse and evolving domain-specific needs are present. \textbf{(4) Suboptimal Strategies for Transfer Learning:} Although recent works have combined SAM with meta-learning for transfer learning, these methods \cite{liu2023matcher,leng2024self} primarily focus on fine-tuning the SAM encoder, implementing teacher-student frameworks through knowledge distillation, or using feature matching by pairwise distance computation \cite{fu2022me}. These approaches often fail to provide comprehensive solutions for efficiently adapting SAM across diverse domains. 
To increase computing efficiency and to better disentangle features of class and domains for model robust learning ability,
%and to achieve high performance,
We propose a task-adaptive visual prompt (TAVP) algorithm that achieves both inter- and intra-domain information disentanglement. In contrast, our method can be effectively generalized to different vertical domains and achieves results comparable to state-of-the-art performance in those domains. The pipeline comparison is shown in Figure \ref{fig: comparison with popular paradigm}.

Upon further analysis of SAM, we observe its poor performance in CD-FSS, which can be attributed to a few key issues. The encoder's image features, though containing basic class data, are mismatched with the target domain's categories, with their inherent distributions potentially causing noise and performance drops. Effective learning requires alignment of feature information with the target domain. Moreover, the decoder's reliance on prompt-based cross-attention mechanisms also hinders its segmentation effectiveness.

Based on the above analysis, we proposed a CDTAP module to better extract class and domain-specific features through contrastive learning from foreground and background, to improve the robustness of CD-FSS, as shown in Figure~\ref{main_fig}. The experimental results show that our work can compute more accurate and robust pairing relationships between samples. Moreover, we propose a fully automatic segmentation framework based on SAM for CD-FSS. This new framework aims to enhance the model's adaptability and accuracy for CD-FSS. Our contributions can be summarized as follows.
\begin{itemize}
    \item We propose a novel model that including a Multi-level Feature Fusion (MFF) and a Class Domain Task-Adaptive Auto-Prompt
(CDTAP) module for efficiently combing SAM with CD-FSS task.
    \item  Compared with SAM, which extracts features in a high-level context, the MFF is proposed to retain low-level feature representations and fuse global and local information to produce class-agnostic features.
\item  To realize the disentanglement of class and domain information, we integrate a unified and comprehensive feature transform method. Specifically, an additional Class-Domain Task-adaptive Auto-prompt (CDTAP) module is proposed for domain and class-specific feature extraction. Simultaneously, We use contrastive learning to achieve deeper and closer matching of samples among different domains.
\item To overcome the shortage of SAM that highly relies on human interaction, we propose an automatic, learnable prompt branch for segmentation, fine-tuning it efficiently with less time and GPU usage, and achieving competitive and best results compared to the state-of-the-art methods on four CD-FSS benchmarks.
\end{itemize}

The overall structure of the paper includes five sections. 
%We introduce the previous related work, especially SAM-based methods, and then propose our new framework for the CD-FSS task in Section~\ref{sec1}. 
We briefly list some related technologies in cross-domain and few-shot learning in Section~\ref{sec1}. The review of related work about cross-domain and few-shot learning is described in Section~\ref{sec2}. We describe our method in more detail and show the model performance in Section~\ref{sec:method}. Then, we conduct comprehensive experiments comparing previous methods and the SAM baseline in Section~\ref{sec4}.
%and ~\ref{sec5}. 
Finally, we conclude this paper and discuss the future application prospects in large model-based fine-tune methods in Section~\ref{sec5}.
\section{Related Works}
\label{sec2}
We start this section by introducing a cross-domain segmentation task with its relative technology, and the few-shot segmentation task is described for related background. Then, we develop the CD-FSS task and the related research within this field.

\subsection{Domain Adaptation in Segmentation}
In recent years, domain adaptation semantic segmentation has made notable progress. To enhance the domain adaptation method, CRTL was proposed by Wang et al. \cite{8891894} by leveraging class priors and a projected Hilbert-Schmidt Independence Criterion (pHSIC) through transfer learning. 
Domain adversarial training is utilized to learn domain-invariant representations in features \cite{tsai2019domain}. Hoffman et al. \cite{hoffman2016fcns} integrated global and local alignment methods with adversarial training.
Other approaches in domain adaptation have also been proposed, such as distillation loss \cite{chen2018road}, output space alignment \cite{tsai2018learning}, class-balanced self-training \cite{zou2018unsupervised}, and conservative loss \cite{zhu2018penalizing}, based on a predefined curriculum learning strategy \cite{zhang2019curriculum}. These methods collectively contributed to advancing adaptive semantic segmentation by leveraging information from various domains, ensuring the model’s robust performance across diverse and less annotated environments. Suppose that the training data originate solely from a single domain and adaptation occurs to an unseen domain. Then, in this case, single-source domain adaptation becomes more challenging due to the limited diversity within the training domain. Consequently, a prevalent approach to address this issue is using data augmentation techniques to generate new domains, thereby enhancing the diversity and information content of the training data. Several methods with different generation strategies were designed to address the single-source domain adaptation problem in computer vision tasks. For example, RandConv \cite{hariharan2011semantic} employed random convolutions for data augmentation. MixStyle integrated style information from instances of randomly selected different domains.

However, previous methods assumed that the target domain shares a similar distribution with the source domain. However, these methods have limitations when applied to scenarios with large distribution gaps, such as medical images and remote sensing images. In contrast with the above data augmentation methods, we use the foundation model to ensure rich prior knowledge instead of generating many images in the source domain, saving computing resources and increasing computation efficiency.
\subsection{Few Shot Segmentation}
Few-shot segmentation (FSS) tasks aim to segment new semantic objects through a limited number of available labelled or unlabeled images that are semantically distinct. Current methods primarily focused on improvements during the meta-learning phase. Prototype-based methods \cite{lee2022rethinking,10138737,}  utilized a technique in which representative foreground or background prototypes were extracted from the support data and various strategies were employed to facilitate interactions either between different prototypes or between prototypes and query features. Relation-based methods \cite{tian2020prior,yang2020prototype,zhang2019pyramid} also succeeded in few-shot segmentation. HSNet \cite{tan2023hsnet} built a high correlation using multi-scale dense matching and captures contextual information using 4D convolution. RePRI \cite{li2021adaptive} introduced transductive inference of base class feature extraction that did not require meta-learning. Besides, to apply the generalization to new classes, Lu et al. \cite{10145054} proposed a Prediction Calibration Network (PCN) for Generalized Few-shot Semantic Segmentation (GFSS), which used a Transformer-based calibration module and cross-attention to reduce class bias and improve segmentation. Chen et al. Moreover, \cite{10436544} proposed a dual-branch learning method for few-shot semantic segmentation, addressing intra-class and inter-class challenges by enhancing feature representations and generalizability to novel classes.
However, these methods primarily focused on segmenting new categories from the same domain with computationally intensive similarity calculation. Due to the significant differences in cross-domain distributions, they failed to be extended to unseen domains.

In contrast to the previous computing prototypes from the class level, we propose a Foreground and Background dual prototype matching method, ensuring fine-grained and class-domain agnostic feature representation.

\subsection{Cross-domain Few-shot Segmentation}
Existing Cross-Domain Few-Shot Learning (CDFSL) methods aim to generalize models to new domains and unseen classes but typically require access to source domain data during pre-training. To reduce the reliance on source domain data, Xu et al. \cite{10471355} proposed an IM-DCL method for Source-Free Cross-Domain Few-Shot Learning (SF-CDFSL), addressing limited labeled target samples and domain disparities through transductive learning and contrastive learning. However, when dealing with more fine-grained tasks such as segmentation, traditional methods often underperform. In this context, Cross-Domain Few-Shot Segmentation (CD-FSS) has emerged as a specialized area, addressing these challenges with benchmarks and novel strategies.

There are four benchmarks \cite{lei2022cross} available for CD-FSS standard evaluation. For the ChestX dataset, the image format has been changed from RGB to gray, with a large gap from the original domain. The other two datasets have more edge information requiring high-quality semantic segmentation.
RD \cite{wang2022remember} introduced a novel domain enhancement strategy leveraging a memory mechanism. This approach involved continuously storing domain-style information from the source domain during the training phase. Subsequently, during testing, this stored source information was utilized to enhance the segmentation performance. During testing, source domain information stored in memory was loaded for the target domain feature enhancement. RD \cite{wang2022remember} offered a direct approach to reduce domain differences and was validated on typical partitioned datasets. For semantic segmentation tasks in autonomous driving applications, PixDA \cite{tavera2022pixel} introduced an innovative pixel-by-pixel domain adversarial loss based on three key criteria: (i) aligning the source and target domains for each pixel, (ii) preventing negative transfer on correctly represented pixels, and (iii) regularizing the training of infrequent classes to mitigate overfitting. CDTF \cite{lu2022cross} achieved cross-domain few-shot segmentation by aligning support and query prototypes. This alignment was realized using an uncertainty-aware contrastive loss and supplemented with a supervised cross-entropy loss and an unsupervised boundary loss as regularization terms. CDTF \cite{lu2022cross} enabled the model to generalize from the base model to the target domain without requiring additional labels. CD-FSS \cite{wang2022cross} presented a cross-domain few-shot segmentation framework that leveraged learning from natural domains to assist in rare-disease skin lesion segmentation. This approach was particularly valuable when dealing with limited data for common diseases in the target domain. PBAL \cite{10102322} came up with a prototype learning and learning technology, which introduced prototype learning and prototype self-training to achieve optimal inter-domain vision and adaptation. However, these methods require a large amount of data for training to achieve a robust model.

Putting aside the previous simple prototype methods, we combine high-level prototype representation with a foundation model, SAM. Besides, we propose a dual prototype matching method for the foreground and background, ensuring fine-grained feature representation.
\subsection{SAM based Methods}
Existing SAM-based segmentation methods incorporate various strategies, including prompt optimization, memory bank feature matching, and adapter modules, to enhance model performance and generalization across domains.
Decoupled SAM (DeSAM) is proposed to address the domain shift issue in medical image segmentation \cite{gao2024desam}. It introduces a prompt-relevant IoU module (PRIM) and a prompt-decoupled mask module (PDMM) to reduce performance degradation caused by poor prompts, achieving enhanced cross-domain robustness on prostate and abdominal datasets.
A source domain prior-assisted module is proposed to enhance the generalization of SAM-based medical image segmentation across domains \cite{dong2024source}. By utilizing a memory bank to store source domain features, the model matches target domain features with these priors to adapt and improve segmentation accuracy.
The CDSG-SAM pipeline is proposed to improve cross-domain few-shot brain tumor segmentation, integrating SAM with a Cross-domain Self-attention (CDS) Adapter and a Self-Generating (SG) Prompt module \cite{yang2025cdsg}.
\section{Methology}
\label{sec:method}
In this section, we introduce our proposed framework TAVP. First, we describe the problem definition. Then, we present an overview of the proposed approach, followed by a detailed explanation of each technical component of the method.

\subsection{Problem Definition}
In the field of cross-domain few-shot semantic segmentation (CD-FSS), let $X_s$ and $X_t$ stand for the input distributions in the source and the target domains, respectively, and $Y_s$ and $Y_t$ denote the label spaces in the two domains, respectively. 
We distinguish between $\{X_{s}, Y_{s}\}$ and $\{X_{t}, Y_{t}\}$ with differing input distributions and non-overlapping label spaces, i.e., $X_{s} \ne X_{t}$ and $Y_{s} \cap Y_{t} = \varnothing$. 
%Here, $X$ signifies the input distribution and $Y$ denotes the label space. 
Our methodology involves training and evaluating our model episodically within a meta-learning framework as outlined in \cite{lei2022cross}. Training episodes consist of a support set and a query set. The support set $S = \{(I_i^s, M_i^s)|i=1, 2, \cdots, K\}$,
%where $i=1$ to $K$, 
where $I_i^s$ is the $i$-th support image and $M_i^s$ is the respective binary mask. 
%with $I$ representing the $i^{th}$ support image and $M_i^s$ is the matching binary mask. 
The query set, defined by $Q = \{(I_i^q, M_i^q)|i=1, 2, \cdots, K\}$, operates similarly. The model is fed with the support set $S$ and a query image set $I^q=\{I_i^q\}_{i=1}^K$, from a specific class $c$, upon which the binary mask set $M^q=\{M_i^q\}_{i=1}^K$ is predicted. 
\subsection{Method Overview}
While SAM can be generalized to more scenarios, even zero-shot situations, it still has some limitations. Firstly, the original SAM relies on interactive prompts for accurate segmentation in different situations, which can be time-consuming. The second challenge is how to transfer richer knowledge and key information from LVM methods while maintaining strong generalization ability. To address these two challenges, we propose an automated framework for segmentation that uses automatic prompts instead of user-interactive prompts. Additionally, we have designed an extra branch for class- and domain-agnostic feature extraction and task-adaptive prompt generation.

\begin{figure*}[!h]
\includegraphics[width=7.3in, height=4.3in]{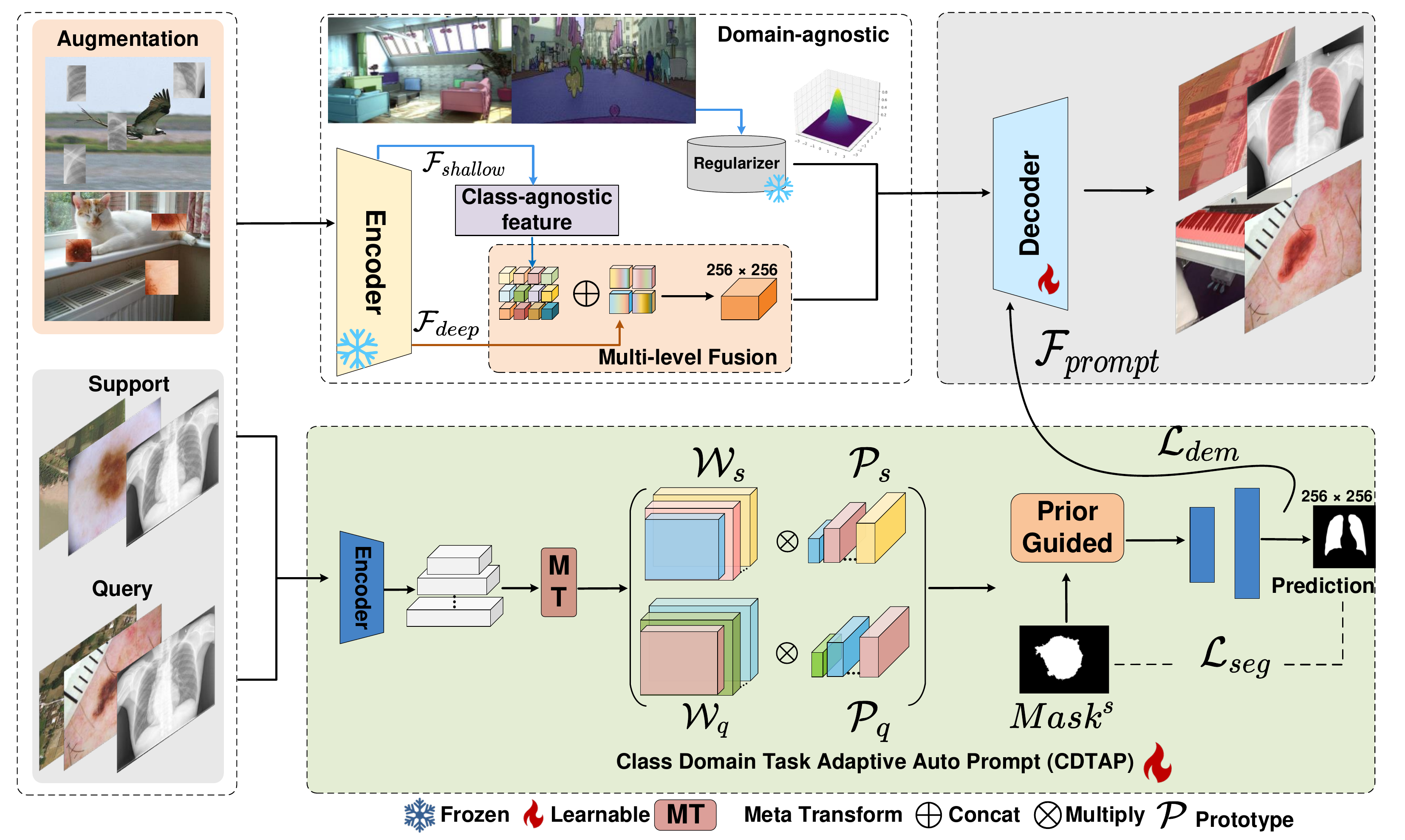}
\caption{The overall architecture of proposed TAVP network. First, the images from source domain with cut-mix are passed through the SAM encoder to obtain multi-level features, which are combined with original pre-trained weights on SA-1B dataset \cite{kirillov2023segment}, and followed by a batch normalization layer to get the class-agnostic features. Additionally, CDTAP is employed for fine-tuning and meta-transformation. Simultaneously, dense embedding: $F_{prompt}$ and image embeddings are acquired as the input of the decoder. At last, the mask decoder predicts the query image. The $L_{dem}$ loss is used for learnable prompt supervision and fine-tuning, and $L_{seg}$ is used for supervising auto-prompts generation.}
\label{main_fig}
\end{figure*}

The overall framework for TAVP is shown in Figure~\ref{main_fig}. The inputs of images from the source domain with cut-mix are fed into the SAM encoder for basic feature extraction. Note that we propose a multi-level feature fusion for extensive representation. Meanwhile, one pair of support and query images from the target domain are fed into the CDTAP module for class domain-specific and agnostic feature extraction. At the same time, this module generates learnable prompts as dense embedding input to the decoder. Then, the combined multi-level and dense prompts are fed into the SAM decoder for prediction.

\subsection{Multi-level Features Fusion}
\textbf{High-level Global Feature Representation.}
We propose an advanced approach to enhance the mask resolution in SAM by incorporating efficient token learning. Rather than utilizing the coarse masks generated by SAM directly, our method involves a High-level token alongside a novel mask prediction layer to produce higher-quality masks.
In this method, we maintain the original mask decoder of SAM but augment it with a newly defined learnable High-level token with the size of $1\times 256$. This token is combined with the existing output tokens with the size of ($4\times 256$) from SAM and prompt tokens with the size of $N_{prompt}\times 256$), serving as the augmented input for the SAM mask decoder. Like the original output token function, the High-level token engages in self-attention with the other tokens participating in token-to-image and image-to-token attention processes within each attention layer for feature refinement. The High-level token utilizes a shared point-wise MLP across decoder layers. After two decoder layers, it comprehensively understands global image semantics and conceals mask information from other output tokens. A novel three-layer MLP is then employed to derive dynamic convolutional kernels from the enriched High-level Token, executing a spatial point-wise operation with the amalgamated High-level feature to generate superior-quality masks.

Our approach trains only the High-level token and its associated three-layer MLPs to correct inaccuracies in the mask produced by SAM without directly fine-tuning SAM or using a post-refinement network. This method stands in contrast to traditional approaches in high-quality segmentation models. Our extensive testing highlights two primary benefits of this efficient token-learning technique. First, it substantially elevates the mask quality of SAM with only a minimal increase in parameters, thus optimizing the training process in terms of time and data efficiency. Second, adaptive token and MLP components prevent overfitting, preserving SAM's zero-shot segmentation performance on new images without knowledge loss.

\textbf{Global and Local Feature Fusion.}
Accurate segmentation requires input features with global semantic context and precise local boundaries. To enhance mask quality further, we augment the mask decoder features of SAM with both advanced object context and refined edge information. Rather than directly utilizing the mask decoder feature of SAM, we construct new multi-level features by extracting and integrating features from various stages of the SAM model. We first extract detailed low-level edge information from the initial layer's local feature of SAM's ViT encoder with a spatial dimension of $64\times 64$. This feature is obtained from the first global attention block within the ViT encoder, specifically the 6th block of 24 blocks in the case of a SAM based on ViT-Large. Then, the last layer's high-level global feature from SAM’s ViT encoder, sized at $64\times 64$, provides a comprehensive global image context. Finally, the mask feature within SAM’s mask decoder, sized at $256\times 256$, is shared by the output tokens and possesses strong shape information of the masks.
As depicted in Figure~\ref{main_fig}, we initially upsample the early and final layer encoder features to a spatial size of $256\times 256$ via transposed convolution to generate the input high-level features. Following this, we combine these three types of features through element-wise summation after straightforward convolutional processing. This approach of fusing global and local features is straightforward yet effective, producing segmentation results that preserve detail with minimal memory and computational costs. In the experimental section, we further conduct a detailed ablation study to assess the impact of each feature source.

\subsection{Class Domain Task Adaptive Auto Prompt (CDTAP)}
We improve the model's generalization by disentangling class-domain prototype information and using prior-guided prompts for fully automatic, task-adaptive prompt generation. The learnable prompt embedding increases the robustness of SAM and our model.
% \begin{figure}[!h]
% \centering
% \hspace{-0.2cm}\includegraphics[width=3.4in, height=1.62in]{figs/DUAL.pdf}
% \caption{ (a)The previous class-wise few-shot methods. (b)Our two-way matching meta-learning module.}
% \label{dual-matching}
% \end{figure}
\begin{figure}[!h]
\centering
\includegraphics[width=3.6in,height=1.4in]{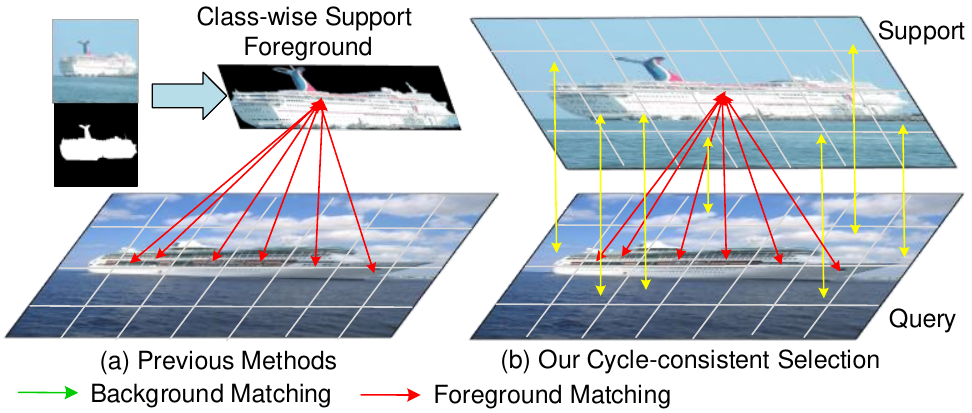}
    \caption{Existing class-wise few-shot methods and our two-way matching meta-learning module.}
    \label{dual-matching}
\end{figure}

\begin{figure}
    \centering
    \includegraphics[width=3.5in, height=1.1in]{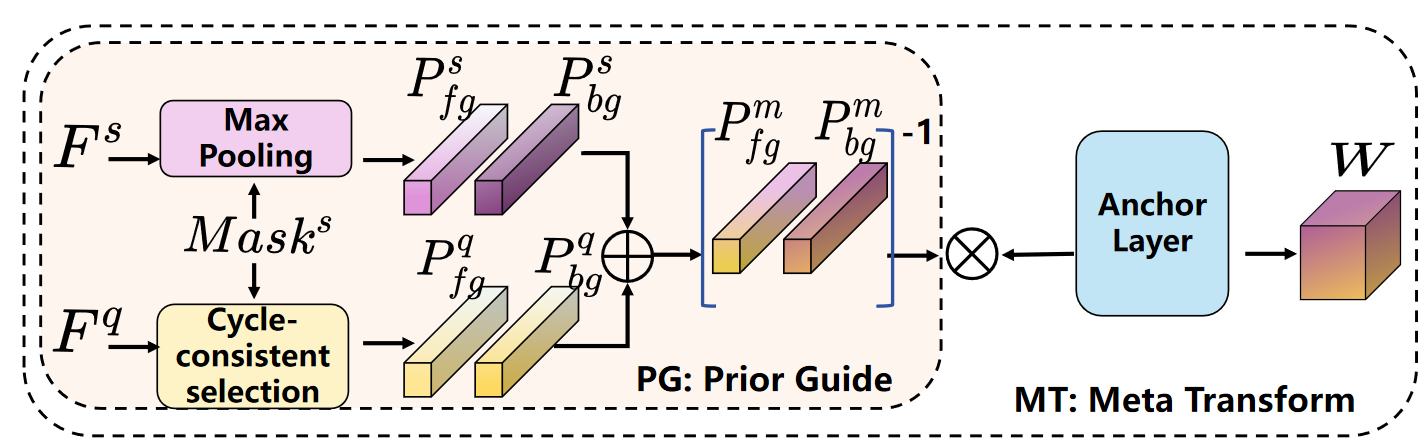}
    \caption{Details of MT and PG.}
    \label{cycle}
\end{figure}

\textbf{Class Domain Prototype Information Disentanglement.}
Previous meta-learning methods only have generalizations for new categories, but the performance degrades when handling both cross-domain and few-shot tasks. To address this, we propose a new prototype-based class-domain information disentanglement module. This module aims to better explore the correlation between class and domain features by separating them into class-domain-common and class-domain-specific components. The foundation segmentation model \cite{kirillov2023segment} is used for base knowledge regulation and a branch for the foreground and background prototype calculations is added. Pixel-level prototype calculations fully utilize feature representations, which is beneficial for few-shot learning.

The pre-trained foundation segmentation model contains a large number of base class knowledge. We extract low-level features for wider expression and high-level features for semantic expression. Then, a batch normalize layer is used for regularization to get class-agnostic knowledge. 

Previous methods only rely on the support prototype set and anchor layer to calculate the transformation matrix. Due to intra-class variance, the support prototype cannot represent all the information in the category. Therefore, we propose to enhance the set of supported prototypes by querying the prototypes. We specifically focus on dual prototype enhancement and cross-domain feature transformation. We leverage cycle consistency between support and query functions to obtain query foreground and background prototypes. Based on these enhanced prototypes that can represent categories and their surroundings, learnable domain-agnostic modules can be used to compute efficient transformation matrices. The transformation matrix is then applied to the query features for cross-domain feature transformation.
Representational archetypes are important for our cross-domain transformation. To this end, we construct a pixel-level fine-grained self-cycling supervision that reasons the query foreground and background to support enhancement. We perform forward matching to obtain the query features with the highest similarity to the supporting prospects. We then use these identified forward-matching query features to backward retarget the corresponding supporting features. If the supporting features found by reverse matching fall within the true supporting foreground mask, the identified query features are averaged and used to derive the foreground prototype. An enhanced background prototype is obtained through the same process. Let $W$ represent the weight matrix of original features in the Anchor layer and $P$ represent the prototype of the foreground representation. Specifically, we use $WP=A$ to obtain $P$ from $W$ and $A$ \text{\cite{lei2022cross}}, where $W$ is a learnable weight matrix, $P$ is the computed prototypes and $A$ is a representation matrix calculated from the distance between the center and other features. The difference between our algorithm and HQ-SAM \cite{ke2024segment} for Multi-level features lies in the design of the CDTAP module, which transfers the original multi-scale based on feature computation to multi-scale based on prototype computation.
The prototype of foreground and background can be calculated by Equation ~\ref{Eq.1}
\begin{equation}
    P_{f,b} = \begin{bmatrix} \frac{P_f}{||P_f||},& \frac{P_b}{||P_b||}\end{bmatrix},
    \label{Eq.1}
\end{equation} 
\begin{equation}
    i^{s->q}=\text{argmax}(\text{sim}(P^s_f\odot P_m^s),P_f^q),
\label{Eq.2}
\end{equation}
\begin{equation}
    j^{q->s}=\text{argmax}(\text{sim}(P^{q,f}_{i^{s>q}},P^{s,f}_{j^{q->s}})),
\label{Eq.3}
\end{equation}
where $i$ and $j$ are rows and columns of 2D spatial positions of the feature map. Equation~\ref{Eq.2} and Equation~\ref{Eq.3} are the cycling check process, where $P^s_f$ is the prototype representation of an image, $P_m^s$ is the prototype representation of its mask, $\odot$ represents the multiplication between vectors, and $P^{q,f}_{i^{s->q}}$ is the feature prototype of the query from support to query matching. The corresponding interpretation can be deduced for $P^{s,f}_{j^{q->s}}$. Given the based equations, the prototype representation of the query can be selected.

We perform class-domain information disentanglement by completing class-domain agnostic feature transformation. Thus, in this branch, we can compute the transformation matrix $A$ for input by calculating its foreground prototype given its corresponding mask in the Anchor layer. 
In $l_{th}$ layer, $m$ represents the mask, $C$ represents the class, $H$ represents its height, and $W$ is the width. The foreground prototype of the support set can be calculated by 
\begin{equation}
p^{l} _{s,f} =\frac{\sum _{i}\sum_{j} f^{l,i,j}_{s,f}\phi ^{l}(m^{i,j}_{s,f})}{\sum _{i}\sum_{j}\phi^{l}(m^{i,j}_{s,f})},
\label{Eq.4}
\end{equation}
where $p^{l}_{s,f}\in R^{C_l}$, $i$ and $j$ are rows and columns of 2D spatial positions of the feature map, $\phi(\cdot)$ denotes a function that bilinearly interpolates input tensor to the spatial size of the feature map $f^{l, i,j}_{s,f}$ at intermediate layer l by expanding along channel dimension, and $\phi(\cdot): R^{H \times W} \to R^{C_l \times H_l \times W_l}$, $m^{i,j}_{s,f}$ is the foreground prototype from support of mask. The support and query sets' background prototypes can be calculated similarly. 

\textbf{Prior Guided Learnable Prompts.}
An essential advantage of SAM is the support of prompt input. However, it is time-consuming for humans to generate interactive prompts, and the decoder of SAM is always coupled with image and prompt embedding. It is reasonable that the prediction can be more accurate with higher-quality prompts. This work proposes the generation of prior-guided meta-space learnable prompts. First, the features are mapped to a new space through the previous two-way enhanced prototype information disentanglement, and the most similar features and their label representations calculated in the query set from the support set are used as prior guides to generate prompts. Then, the enhanced inputs, including multi-level image embeddings with the size of $256 \times 256$ and high-quality prompts of similar size, are fed into a high-quality decoder.

\subsection{Light-Weight Fine-tune Framework}
Besides, we adopt a random heterogenization sampling strategy to distinguish different cross-domain tasks. In this approach, a threshold value is set to monitor the quality of the sampling process. One of the limitations of SAM is that it is time-consuming and inefficient, which is a common issue in large model fine-tuning. In this work, we propose a lightweight fine-tuning framework, transferring SAM to cross-domain few-shot segmentation only by re-training a few layers in CNN-based models. First, the target domain samples are fed into the class domain task-specific branch for class-agnostic feature extraction. These highly structured, class-agnostic feature embeddings, together with other feature embeddings from the base domain, are fed into the decoder. A weighted supervision loss is proposed to fine-tune the decoder to predict masks for target domain samples. \textit{L$_{seg}$} represents the segmentation loss function, composed of the Cross-Entropy loss function \cite{shannon2001mathematical}, and a Dice loss function \cite{milletari2016v}, as defined in 
\begin{equation}
    L_{seg} = (1-\lambda ) \cdot L_{CE} + \lambda \cdot L_{Dice},
\label{Eq.Lseg}
\end{equation}
where $\lambda$ is an adjustable parameter for supervision.
Simultaneously, samples from the target domain are fed into a CNN-based model to generate dense embeddings as auto prompts. The dense embedding is obtained from the layer of CNN-based backbones as a weight matrix aligned with a feature map. Then, the dense embedding is multiplied with a combined multi-level feature map and is fed into a decoder, achieving guided decoding for target domain samples. Given an input $x$, it is fed into a CNN-based encoder. After down-sampling, a simple decoder follows for up-sampling to generate dense embedding, aligning with the feature map. The $L_{dem}$ loss function is adopted to supervise dense embedding:
\begin{equation}
L_{dem}(x) = L_{BCE}(Z_x,M_x) +  L_{Dice}(Z_x,M_x),
\label{Eq.5}
\end{equation}
where $Z_x$ represents the dense embedding of input $x$, and $M_x$ is the mask of input $x$. 
Overall, the end-to-end training framework is supervised by 
the following loss function
\begin{equation}
    L = L_{seg} + L_{dem},
\label{Eq.loss}
\end{equation}
where \textit{L$_{seg}$} represents segmentation branch supervision and \textit{L$_{dem}$} is the loss function for dense embedding generating.

\section{Experiments}
\label{sec4}
In this section, we describe the experimental settings, including `datasets', `Data Pre-processing Strategy', `Models Baseline', and `Implementation Details'.

\subsection{Experimental Settings}
We first introduce the benchmarks in the CD-FSS. Next, the model baseline, implementation details, and performance visualization are listed.
\subsubsection{datasets}
In cross-domain few-shot segmentation, four benchmarks are contributed \cite{lei2022cross}. 

\textbf{Deepglobe.}
The Deepglobe dataset, described in \cite{demir2018deepglobe}, is a collection of satellite images. It includes pixel-level annotations for seven categories: urban areas, agriculture, rangeland, forest, water, barren land, and an `unknown' category. In total, 803 images in the dataset have a consistent spatial resolution of $2448 \times 2448$ pixels.

Follow the standard approach to previous work \cite{lei2022cross}, We divide each image into six sections to increase the number of testing images and reduce their sizes. Since the object categories in this dataset have irregular shapes, cutting the images has minimal impact on their segmentation. We further filter out images with only one class and those belonging to the `unknown' category. This results in 5,666 images used to report the results, each with a resolution of $408 \times 408$ pixels.

\textbf{ISIC}. The dataset identified as `document number 1', as described in  \cite{codella2019skin,tschandl2018ham10000}, focuses on skin lesion imagery from cancer screenings, containing 2,596 images with a single lesion. Ground-truth labels are provided solely for the training set. Follow the standard approach to previous work \cite{lei2022cross}, for consistent analysis, images are resized to a standard $512 \times 512$ pixels from the original $1022 \times 767$ pixels.

\textbf{ChestX.}
As discussed in \cite{jaeger2013automatic,candemir2013lung}, the Chest X-ray dataset is tailored for Tuberculosis detection. It comprises a total of 566 X-ray images, each with an original resolution of $4020 \times 4892$ pixels. These images are sourced from a dataset of 58 cases with a Tuberculosis manifestation and 80 cases with normal conditions. Given the large size of the original images, a common practice is to reduce them to a more manageable $1024 \times 1024$ pixels for further analysis and processing.

\textbf{FS1000.}
FSS-1000 \cite{li2020fss} is a natural image dataset for few-shot segmentation, consisting of 1,000 object classes, each with 10 samples. We use the official split for semantic segmentation in our experiment and report the results on the official testing set, which contains 240 classes and 2,400 testing images.

\subsubsection{Data Augmentation and Sampling Strategy.}
In this work, a cut-mix method and a heterogenization sampling strategy are adopted to reduce the coupling effect of training with a limited dataset. First, images from the target domain are randomly divided into patches of different sizes, and then the image from the original domain is used as the background to create a new input image. In the experiment, each newly synthesized image contains 5 patches from the original domain. In the second strategy, a threshold is set to control the quality of the sampling. Specifically, we applied a 5-fold validation strategy on the split dataset during the training process. The threshold is computed dynamically during training to select samples for model training. The size of each patch is 4892 pixels.

Besides, several data augmentation methods are also used in the original SAM baseline, including adjusting the attributes of images such as brightness, contrast, saturation, etc., randomly flipping along vertical and horizontal levels, and random affine transformation.

After the experiments, we emphasize the importance of ensuring that the CD-FSS task relies on improved guidance and representation of foreground and essential information for the input sample, while highlighting that the background plays a critical role in achieving more accurate predictions. This emphasizes the efficiency of our method as shown in Figure \ref{dual-matching}. For example, the more random the background setting is, or the greater the difference between the data in the source domain is, the better the model performs.
\begin{table*}[ht!]
    \caption{\small Comparison with previous FSS and CD-FSS methods under 1-way 1 shot and 5-shot settings on the CD-FSS benchmark.}
\label{tab1}
 \setlength{\tabcolsep}{3.3mm}{
\begin{tabular}{@{}c@{}|c|c|c|c|c|c|c|c|c|c|c}
\toprule
%\multirow{2}{*}{Methods} & \multirow{2}{*}{Backbone}
Methods & Backbone
& \multicolumn{2}{c|}{ISIC}  & \multicolumn{2}{c|}{Chext X-ray} &  \multicolumn{2}{c|}{Deeepglobe}& \multicolumn{2}{c|}{FSS1000}& \multicolumn{2}{c}{Average}                \\
\midrule
 \multicolumn{2}{c|}{Task}  
 & 1-shot      & 5-shot      & 1-shot         & 5-shot         & 1-shot        & 5-shot    &   1-shot        & 5-shot        &   1-shot        & 5-shot   \\
\midrule
\multicolumn{10}{c}{Few Shot Segmentation Methods}\\
\midrule
AMP \cite{siam2019amp}               & VGG-16                    & 28.42       & 30.41       & 51.23          & 53.04             & 37.61                 & 40.61  & 57.18 & 59.24 &   43.61  &  45.83   \\
PGNet \cite{zhang2019pyramid}              & ResNet-50                 & 21.86      & 21.25       & 33.95          & 27.96          & 10.73                 & 12.36  & 62.42  & 62.74 & 32.24  &   31.08   \\
PANet \cite{wang2019panet}            & ResNet-50                 & 25.29       & 33.99       & 57.75          & 69.31            & 36.55                 & 45.43 & 69.15 & 71.68  &   47.19   & 55.10  \\
CaNet \cite{zhang2019canet}             & ResNet-50                 & 25.16       & 28.22       & 28.35          & 28.62          & 22.32                & 23.07  & 70.67  & 72.03  &  36.63    &  37.99 \\
RPMMs \cite{yang2020prototype}          & ResNet-50                 & 18.02       & 20.04       & 30.11          & 30.82               & 12.99                 & 13.47  & 65.12  & 67.06  & 31.56     &  32.85    \\
PFENet \cite{tian2020prior}          & ResNet-50                 & 23.50       & 23.83       & 27.22          & 27.57              & 16.88                 & 18.01  & 70.87  & 70.52  &  34.62   &  34.98    \\
RePRI \cite{boudiaf2021few}          & ResNet-50                 & 23.27       & 26.23       & 65.08          & 65.48           & 25.03                 & 27.41  & 70.96  & 74.23 &  46.09   &  48.34    \\
HSNet \cite{min2021hypercorrelation}     & ResNet-50                 & 31.20       & 35.10       & 51.88          & 54.36           & 29.65                 & 35.08   & 77.53 & 80.99 & 47.57  &  51.38    \\
\midrule
\multicolumn{10}{c}{ViT Based Methods and Cross Domain Few Shot Segmentation Methods}\\
\midrule
PATNet \cite{lei2022cross}        & ResNet-50                 & 41.16       & 53.58       & 66.61      &70.20    & 37.89            & 42.97  & 78.59  & 81.23   & 56.06   &  61.99 \\
RestNet \cite{huang2023segment}           & ResNet-50                 & 42.25       & 51.10       & 71.43          & 73.69    &35.68 & 39.87 & \textbf{81.53}  & 84.89  &  56.84 & 62.39\\  
$\text{IFA}_{T=3}$ \cite{nie2024cross}   & ResNet-50   & 66.3 & 69.8      & 74.0 &  74.6   &50.6 & 58.8 & 80.1  & 82.4  &  67.8  & 71.4\\ 
APM-M \cite{tong2024lightweight}   & ResNet-50   & 41.71  & 51.16   & 78.25 & 82.81   &40.86 & 44.92  & 79.29  & 81.83   &  60.03 & 65.18\\ 
DMTNet \cite{chen2024cross}   & ResNet-50   & 43.55   & 52.30   & 73.74 & 77.30  &40.14 & 51.17& 81.52 &  83.28 &  59.74 & 66.01\\ 
HQ-SAM \cite{ke2024segment}& ViT   & 40.38  &  47.60    &   62.86       &  73.14  &24.73 &26.82  &78.97 & 80.97  & 51.74 & 57.13\\
SAM-Med2d & ViT   & 62.37  &  65.40    &   65.91       & 70.85  & 16.78&18.58 &73.54 & 76.80  & 54.65  & 57.91\\
SAM-Adapter& ViT   & 33.47  &  38.33    &       53.99   & 58.05   & 45.79&47.65 &67.98 & 70.80  & 50.31  &53.71 \\
\textbf{APSeg} \cite{he2024apseg} & ViT   & 45.43       & 53.89       & \textbf{84.10 }  & 84.50    & 35.94 & 39.98  &  79.71 & 81.90   &  61.30 & 65.09\\
\textbf{TAVP(ours)}   & ViT + ResNet & \textbf{54.89}  & \textbf{73.39} & 70.31& \textbf{88.61}     & \textbf{46.10}    & \textbf{61.98}  & 79.09  & \textbf{83.41} & \textbf{62.60}    &  \textbf{76.85}   \\ 
\bottomrule
\end{tabular}}
\end{table*}
\subsubsection{Implementation Details.}
We have designed three backbones for the additional branch except the SAM framework. SAM encoder is adapted for high-level global semantic contexts and low-level local information extraction. The full ViT is used for global semantic contexts, and the local low-level features are extracted from the early layer. Besides, a CNN-based encoder, modified for computing prototypes, is used for class-domain agnostic feature extraction, and a small up-sampling method is adopted for dense embedding generation as auto-prompts.

There are two parts of input here. The first part uses enhanced data, a cutmix library (a total of 1,000 images) is generated to increase feature diversity. We randomly cropped the images in the target domain into different patches and pasted them on the original domain images. Then the augmented datasets are fed into SAM encoder to extract image embedding for late decoding. For the another part, only one pair of support and query images from the target domain (including ChestX \cite{jaeger2013automatic}, ISIC \cite{codella2019skin}, FSS-1000 \cite{li2020fss} and deepglobe \cite{demir2018deepglobe}) are fed into the CDTAP module. For the training parameters, the baseline is frozen, indicated by the snowflake icon in Figure \ref{main_fig}, and only the CDTAP module is trained.

In the training stage, the number of epochs is set to be between 60 and 150. Our experiments can achieve ideal results if the running time is between 2 and 6 hours on NVIDIA A6000 GPU, depending on the number of epochs and validation datasets. 

\begin{figure*}[t]
\centering
\includegraphics[width=7in]{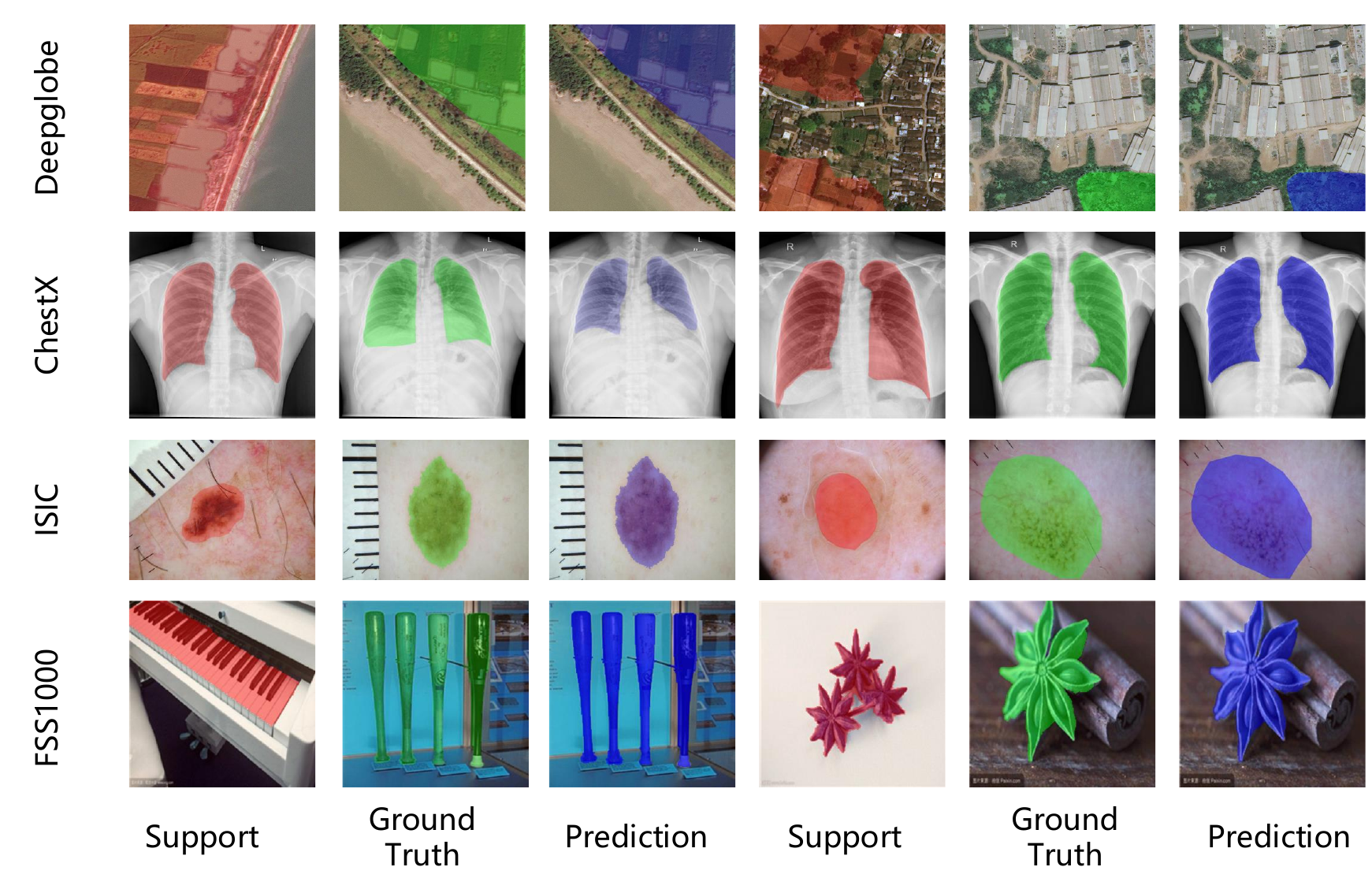}
\caption{\small Qualitative results of TAVP in 1-way 5-shot segmentation on CD-FSS. Support labels are overlaid in red. The ground truth and predictions of query images are highlighted, respectively.}
\label{evaluation}
\end{figure*}

\subsection{Comparison with SOTA Methods}
Extensive experiments are conducted to compare our method with the state-of-the-art methods. The results show that we achieve better results on the Deepglobe dataset than the latest SOTA performance. Besides, in the other three cross-domain datasets, we achieve better and more competitive and accurate results than the previous methods as shown in Table \ref{tab1}. Moreover, it is obvious that with a more robust model and flexible learning ability, the prediction is closer to the samples' original semantics, especially pixel-level information, instead of relying on fixed ground truth, as shown in Figure \ref{evaluation}. 

\subsection{Ablation Study}
Given the proposed method, we test the performance of models with different combination strategies. Overall, we divide the ablation experiment into the following parts based on different backbones, data augmentation strategy, fusion branches, and ablation study with SOTA. All ablation experiments are based on the pre-trained weight: `vit\_h' for better performance. Besides, the FS1000 dataset presents minimal cross-domain difficulty and is not representative, so we do not use it in ablation experiments to test the effectiveness of our algorithms but only use it in comparative experiments in Table \ref{tab1}.

\begin{table}[h!]
\renewcommand{\arraystretch}{1.3}
    \caption{ Ablation Study of different settings under 1-way 1-shot and 5-shot settings.}
    \label{tab2}
 \setlength{\tabcolsep}{1mm}{
\begin{tabular}{l|ll|ll|ll}
\hline
\multirow{2}{*}{Backbone} & \multicolumn{2}{l|}{ChestX}          & \multicolumn{2}{l|}{ISIC}            & \multicolumn{2}{l}{Deepglobe}        \\ \cline{2-7} 
                          & \multicolumn{1}{l|}{1-shot} & 5-shot & \multicolumn{1}{l|}{1-shot} & 5-shot & \multicolumn{1}{l|}{1-shot} & 5-shot \\ \hline
\multicolumn{7}{c}{ResNet50}  \\ \hline 
w/o Data Augmentation     & \multicolumn{1}{l|}{60.14}  & 75.68  & \multicolumn{1}{l|}{21.19}  & 27.32  & \multicolumn{1}{l|}{43.29}  & 53.10  \\ \hline
with Data Augmentation    & \multicolumn{1}{l|}{70.31}  & 88.61  & \multicolumn{1}{l|}{54.89}  & 73.39  & \multicolumn{1}{l|}{46.10}  & 61.98  \\ \hline
\multicolumn{7}{c}{HardNet85}  \\ \hline 
w/o Data Augmentation     & \multicolumn{1}{l|}{61.30}  & 73.70  & \multicolumn{1}{l|}{23.40}  & 31.72  & \multicolumn{1}{l|}{40.98}  & 59.73  \\ \hline
with Data Augmentation    & \multicolumn{1}{l|}{65.79}  & 86.54  & \multicolumn{1}{l|}{56.11}  & 69.79  & \multicolumn{1}{l|}{46.63}  & 56.37  \\ \hline
\end{tabular}}
\end{table}

\textbf{Backbone and Data Augmentation Ablation.}
In the additional task-specific class-domain agnostic feature extraction and auto-prompts generation branch, we perform ablation experiments on the models' performance on three datasets that are more difficult for cross-domain challenges under 1-way 1-shot and 5-shot settings, and the details are shown in Table \ref{tab2}. This study proves that ResNet has a stronger recognition ability for categories, and its effect is outstanding in our novel learnable prompt.

\textbf{Branch Ablation.}
In branch ablation testing based on different backbones, we chose ResNet as the backbone of CNN-based feature extraction in CDTAP. First, MFF is a multi-level feature fusion module. CDTAP is a task-adaptive information disentanglement module. 
We performed ablation experiments with our method in the same setting as previous work of APSeg \cite{he2024apseg} and HQ-SAM \cite{ke2024segment}, and the results are description in Table \ref{tab3}.
\begin{table}[htbp]
 \caption{\small Ablation Study of MFF and CDTAP in the 5-shot setting.}
 \label{tab3}
\renewcommand{\arraystretch}{1.3}
 \setlength{\tabcolsep}{0.05mm}{
\begin{tabular}{l|c|c|c|c|c|c|c|c}
\hline
  \multirow{2}{*}{Model}& \multirow{2}{*}{Backbone} & \multicolumn{2}{c|}{Modules}&  \multicolumn{5}{c}{mIOU(\%)}  \\ \cline{3-9}
      &  & MFF & CDTAP & ChestX & ISIC  & Deepglobe & FSS1000 & Average \\ \hline
 \multicolumn{9}{c}{SAM Baseline}  \\ \hline 
SAM    & ViT        & \ding{55}   & \checkmark     & 43.80  & 50.55 & 23.19     & 78.90   & 49.11   \\ \hline
APSeg  & ViT & \ding{55}   &  \checkmark   & 84.50  & 59.89 &  46.98   & \textbf{81.90}   & 68.32  \\ \hline
Our  & ViT + ResNet  & \ding{55}   & \checkmark    & \textbf{85.01}  & \textbf{60.30} & \textbf {56.98}   & 79.87  & \textbf{70.54} \\ \hline
\multicolumn{9}{c}{SAM ++}  \\ \hline
HQ-SAM & ViT & \checkmark   & \ding{55}     & 30.14  & 47.60 & 26.02     & 80.97   & 46.18   \\ \hline
APSeg  & ViT & \checkmark   &  \ding{55}    & 86.91  & 71.14 & 47.63     & \textbf{83.41 }  & 72.27  \\ \hline
Our  & ViT + ResNet  & \checkmark   & \ding{55}     & \textbf{87.03}  & \textbf{73.39} & \textbf{60.98}     & 82.90   & \textbf{76.08 }  \\ \hline
\end{tabular}}
\end{table}

\textbf{Comparison with SOTA under the Same Setting.}
To provide a fair baseline, we train PATNet with ViT-base, SAM initialization, and $1024 \times 1024$ crops. Results in Table~\ref{my-label} demonstrate again the superiority of our TAVP compared with PATNet under the same settings. We can see that the ViT-based PATNet improves the performance compared to CNN-based PATNet \cite{lei2022cross} on the dataset of Chest X-ray, ISIC, and Deepglobe. These results prove again that these three datasets are more challenging, and our method performs more robustly.
\begin{table}[!h]
    \caption{\small Ablation Study of SOTA under the same setting in the 1-shot scenario.}
\label{my-label}
\renewcommand{\arraystretch}{1.3}
\centering
\setlength{\tabcolsep}{2mm}{
\begin{tabular}{l|l|l|l|l|l}
\hline
 Method & Backbone & Size  & ChestX & ISIC  & Deepglobe \\ \hline
TAVP           & ViT-base          & 1024 x 1024 & 70.31          & \textbf{54.89}    & \textbf{46.10}          \\ \hline
PATNet          & ViT-base          & 1024 x 1024 & 76.43          & 44.25            & 22.37           \\ \hline
\end{tabular}}
\end{table}

\subsection{Efficiency Comparison}
Considering the huge amount of parameter calculation required for the basic model, we only train some fine-tunable parameters. SAM needs to train a model with a large number of parameters from scratch, while our framework only needs to fine-tune some layers and parameters instead of starting from scratch. In addition, linear computation is incorporated into our framework to reduce the number of parameters, thus requiring significantly fewer parameters.
\begin{table}[htbp]
\centering
\caption{\small Ablation Study of efficiency.}
\label{tab:table1}
\renewcommand\arraystretch{1.2}
\setlength{\tabcolsep}{3.8mm}{
\begin{tabular}{c|c|c}
\hline
\textbf{Backbone} & \textbf{Vision Encoders} & \textbf{\#Params(M)} \\ \hline
Hardnet           & CNN                      & 41.56                \\
Hardnet + attention & CNN                      & 46.14                \\ \hline
ResNet  + attention  & CNN         & 38.54                \\
\textbf{CDTAP Module(ours)}              & CNN        & \textbf{36.5}                 \\ \hline
SAM               & ViT-B       & 93.7                 \\
SAM               & ViT-L     & 312.3                \\
SAM               & ViT-H        & 641.1                \\ \hline
\end{tabular}}
\end{table}
\begin{table}[!h]
\centering
\caption{Efficiency Comparision between SAM based models}
\label{tb2}
 \setlength{\tabcolsep}{4.4mm}{
\begin{tabular}{l|l|l|l}
\hline
Method    & Resolution  & \begin{tabular}[c]{@{}l@{}}Learnable \\ Parameters(M)\end{tabular} & FPS \\ \hline
TAVP(ours) & 1024 × 1024 & 36.5                                                               & 12  \\ \hline
SAM       & 1024 × 1024 & 1191                                                               & 8   \\ \hline
\end{tabular}}
\end{table}
The detailed comparison results of efficiency are shown in the Table \ref{tb2}. Table~\ref{tab:table1} shows the detailed parameter comparison. Notice that the bottom three lines are the parameters of the original SAM based on ViT. The other lines are our whole framework's parameters based on different backbones, all smaller than SAM. These results prove again that our method improves the efficiency of SAM and is more light. 
\subsection{T-SNE Visualization}
We use t-SNE plots to visualize the distribution of test data from a specific target domain. In Figure 5 (a), the original distribution shows that the foreground-boundary (F-B) and background-boundary (B-B) pixels are mixed together, making it difficult to clearly differentiate between the foreground and background regions. In contrast, (b) demonstrates the effect of applying a prototype-based clustering method, which reorganizes the data and allows for a more distinct separation of foreground and background information. This shows how t-SNE visualization, combined with prototype-based clustering, enhances the clarity and structure of the data distribution.
\begin{figure}[!h]
    \centering
\hspace{-0.6in}\includegraphics[width=4in, height=1.6in]{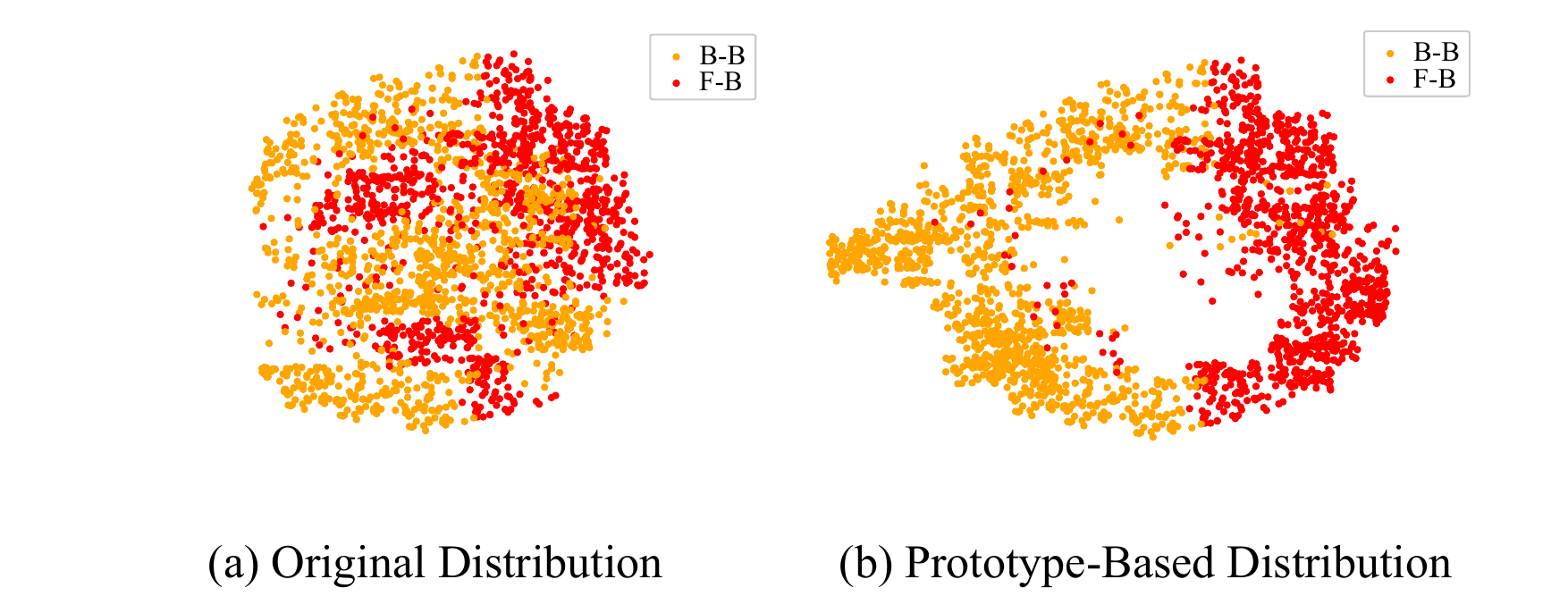}
    \caption{Visualization of foreground-boundary (F-B) and background-boundary (B-B) pixels in t-SNE plots. (a) The original distribution showing the F-B and B-B pixels. (b) The prototype-based distribution displaying the F-B and B-B pixels based on prototype-based clustering.}
    \label{fig:tsne}
\end{figure}
\begin{figure*}[h!]
\begin{center}
\centering
\includegraphics[width=6.7in,height=0.65\textwidth]{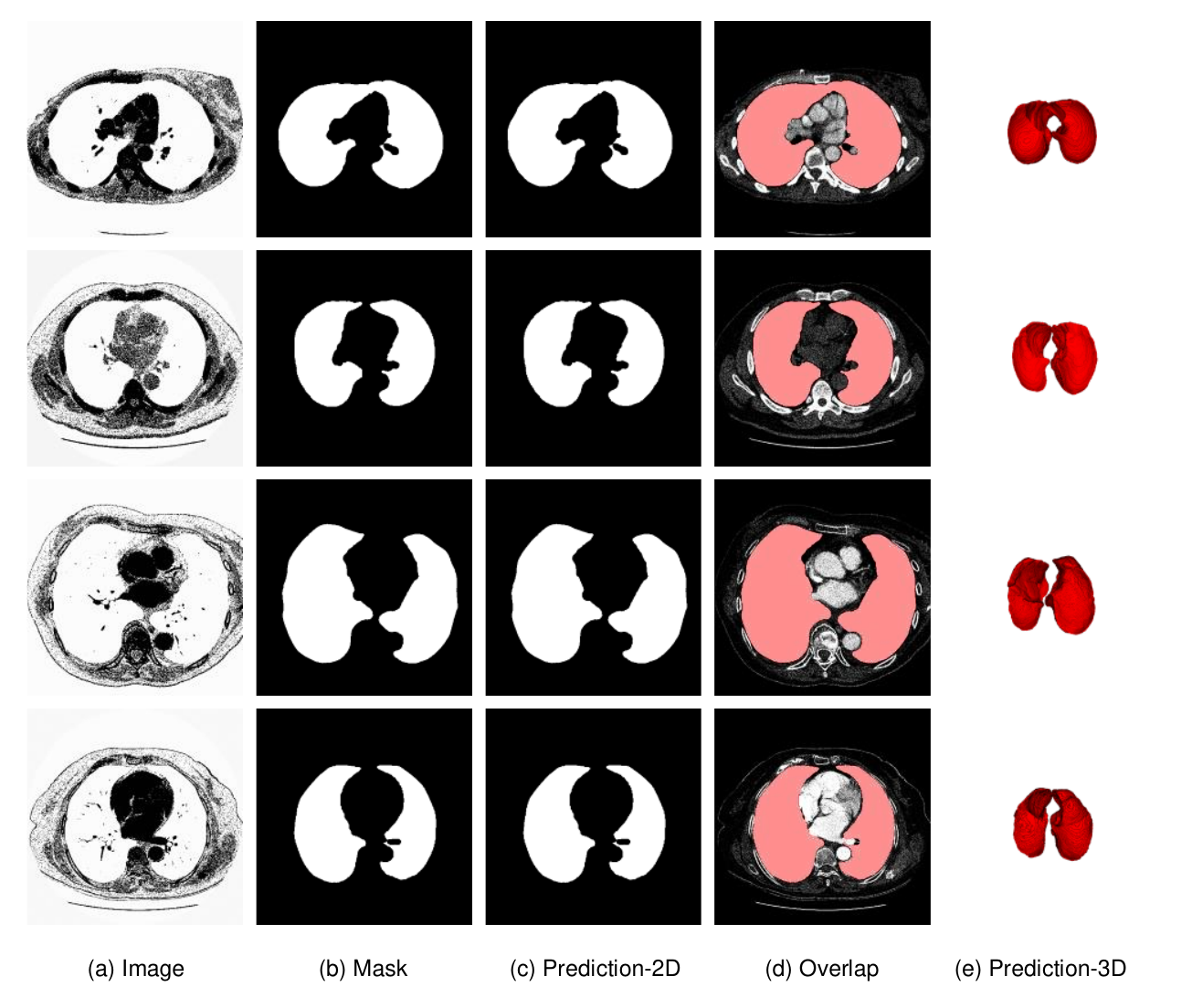}
    \end{center}
    \caption{Performance of our model on a medical image dataset: Lung-CT, the 2d-predictions is shown in (c). We test the 3-dimensional data and print the test mask using a visualization tool, as shown in (e).}
    \label{fig: lung_2d}
\end{figure*}
\subsection{Performance on other datasets}
Except for the benchmarks we compared on the paper, we also tested on other few-shot and cross-domain datasets including CT-Lung, and SUIM.

\textbf{Performance on CT-Lung \cite{jiang2021deep}}
The dataset derived from the Lung Nodule Analysis (LUNA) competition is a collection of CT scans focused on the lungs. The previous study \cite{yang2023dual} used this dataset to test cross-domain performance. It encompasses 534 CT images of the lung, each with a resolution of 512 by 512 pixels. Notably, this dataset is exclusively dedicated to lung-related imagery, and all the images within it are in grayscale.
We tested the model combining CDFS branch and auto-prompt branch, the visualization of performance is shown in Figure ~\ref{fig: lung_2d}.
\begin{figure*}[h!]
\begin{center}
\centering
\includegraphics[width=7.15in,height=0.53\textwidth]{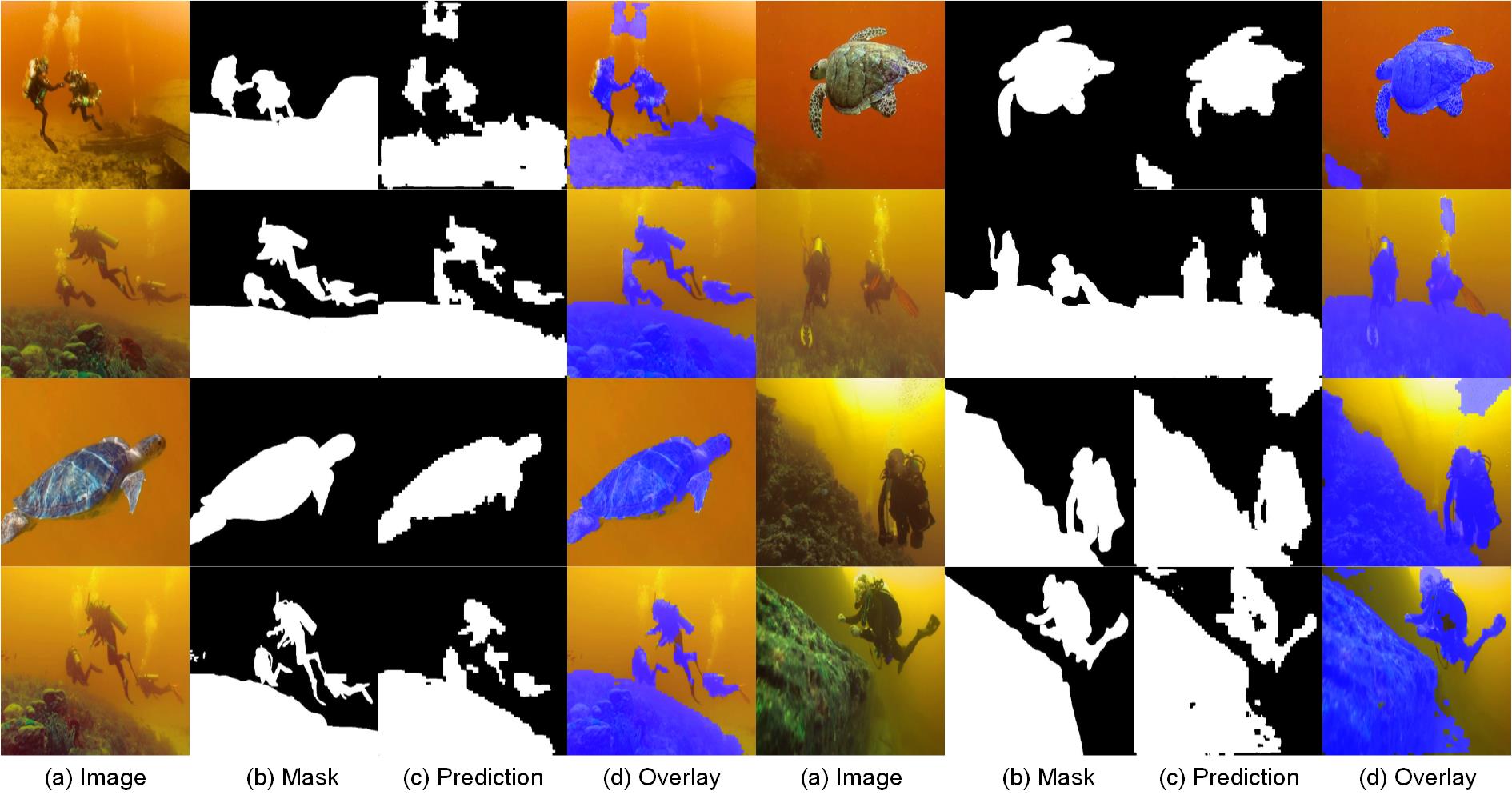}
    \end{center}
    \caption{Performance of our model on an underwater image dataset: SUIM, the predictions are shown in (c) and the overlay is visualized in blue color.}
    \label{fig: SUIM}
\end{figure*}

\textbf{Performance on SUIM \cite{islam2020semantic}}
The SUIM dataset is a specialized resource designed for underwater image segmentation. It comprises a collection of 1,525 images, featuring eight distinct classes: fish, reefs, aquatic plants, wrecks/ruins, human divers, robots, and sea-floor. The dataset is tailored to enhance the performance and accuracy of image segmentation algorithms in underwater environments, a challenging domain due to factors like varying light conditions and water turbidity. By offering a variety of underwater scenes and objects, SUIM plays a crucial role in advancing computer vision technologies for marine research, underwater robotics, and environmental monitoring.
We tested the model combining CDFS branch and auto-prompt branch, the visualization of performance is shown in Figure ~\ref{fig: SUIM}.

\section{Conclusion }
\label{sec5}
%\label{sec:conclusion}
It is worth noting that our work is the first one to apply a large foundation model-based method to CD-FSS tasks, shifting focus from traditional CNN-based deep learning approaches. By leveraging the Segment Anything Model (SAM), a powerful foundational model for segmentation, the proposed framework redefines SAM's role in CD-FSS tasks and introduces a novel perspective on using large models to address domain-specific challenges. The incorporation of the CDTAP module, which enables adaptive and learnable visual prompts, allows for enhanced segmentation accuracy and robustness, achieving state-of-the-art performance on three widely-used CD-FSS benchmarks.

The extensive experiments conducted demonstrate that SAM provides satisfactory results for a variety of segmentation tasks, showcasing its generalization capability. However, the study also highlights limitations in certain scenarios, such as the DeepGlobe dataset, where SAM's performance does not meet expectations, underlining the necessity for further refinement of SAM-based methods to enhance their adaptability and effectiveness in more challenging environments. The proposed framework thus serves as a significant step forward, offering an innovative and efficient pathway for large model transfer in CD-FSS tasks. Beyond practical outcomes, this work opens a new frontier in leveraging foundational models for cross-domain and few-shot learning. SAM’s ability to act as a foundational knowledge tool, transferring its learned representations to new and diverse tasks, is a noteworthy achievement. The adaptive visual prompts introduced in this study provide a flexible mechanism for knowledge transfer, demonstrating the potential for SAM-based approaches to tackle domain-specific segmentation problems effectively.

Moreover, this work serves as an initial exploration into the transfer of SAM’s knowledge through adaptive visual prompts, emphasizing the need for future research into more efficient algorithms with strong learning capabilities. Such advancements will not only enhance CD-FSS performance but also contribute to the broader goal of advancing Artificial General Intelligence by enabling more robust domain adaptation and few-shot learning methodologies.

%\begin{thebibliography}{1}
\bibliographystyle{IEEEtran}
\bibliography{main}
%\end{thebibliography}

\end{document}